\pdfoutput=1
\documentclass{SSBAtran}

\usepackage{graphicx}
\begin{document}
%
\title{A multitask deep learning model for real-time deployment in embedded systems}

\author{\IEEEauthorblockN{%
Miquel Mart\'i\IEEEauthorrefmark{1}\IEEEauthorrefmark{2} and
Atsuto Maki\IEEEauthorrefmark{2}}
\IEEEauthorblockA{\IEEEauthorrefmark{1}Universitat Polit\`ecnica de Catalunya, Barcelona, Catalonia/Spain}
\IEEEauthorblockA{\IEEEauthorrefmark{2}KTH Royal Institute of Technology, Stockholm, Sweden\\
Email: $\{$miquelmr,atsuto$\}$@kth.se}
}

\setlength{\textfloatsep}{10pt plus 1.0pt minus 2.0pt}

\maketitle

\begin{abstract}
We propose an approach to Multitask Learning (MTL) to make deep learning models faster and lighter for applications in which multiple tasks need to be solved simultaneously, which is particularly useful in embedded, real-time systems.
We develop a multitask model for both Object Detection and Semantic Segmentation and analyze the challenges that appear during its training. Our multitask network is 1.6x faster, lighter and uses less memory than deploying the single-task models in parallel.
We conclude that MTL has the potential to give superior performance in exchange of a more complex training process that introduces challenges not present in single-task models.
\end{abstract}


\section{Introduction}
Deep learning models (and machine learning models in general) focus on solving one task at a time. However, applications often require more than one task to be performed simultaneously and the naive solution is to deploy in parallel one model for each task.

Applications imposing real-time constraints require small inference times and prohibit off-board computation, forcing the deployment of deep learning models in embedded systems, in which not only storage and memory available are limited but also computing power. This presents challenges in terms of resources, accentuated when deploying multiple models: weights storage and forward pass memory usage and computational complexity.

Multitask Learning \cite{Caruana} learns to solve tasks in parallel using a shared representation of a common input, improving the generalization capabilities of the models. Multitask networks share a base trunk and a number of task-specific branches emerging from it. Only the task-specific layers are computed separately.

In this work, we propose using multitask models to get benefits in terms of speed, memory usage and storage during deployment. For studying our approach, we train and evaluate a multitask model for both semantic segmentation and object detection. We highlight the challenges imposed by applying MTL, explain how they affect the performance of our model and show that it compares positively in terms of inference time, memory usage and model size against deploying one model per task.

\section{Related Work}
\subsection{Multitask Learning}

MTL has been succesfully used in different domains, including CV \cite{UberNet,MaskRCNN}. Some challenges appear when applying it \cite{Caruana}: \textit{learning speed} differences between tasks and  deciding \textit{what to share} according to the \textit{relatedness} between tasks in the multitask architecture \cite{Stitch, AdaptiveFeatureSharing}.

\subsection{Semantic Segmentation}

Semantic segmentation aims at partitioning parts of images belonging to the same semantic class, typically via pixel-wise classification. Fully convolutional networks (FCN) \cite{FCN} have improved both accuracy and speed for dense prediction problems by using only convolutional layers. Upsampling layers allow a segmentation output size equal to the input and skip connections add finer details. Other approaches add post-processing steps \cite{DeeplabCRF}, learnable \textit{deconvolution} layers \cite{ Deconv} or global context \cite{ParseNet}.

\subsection{Object Detection}

Object detection aims at finding in an image all instances of objects and classifying them in a number of classes. Faster R-CNN \cite{FasterRCNN} was the first to give close to real-time performance. YOLO \cite{YOLO} avoids the generation of region proposals for increased speed. SSD \cite{SSD} avoids fully-connected layers for speed and takes features at different levels for improved accuracy. 


\section{Methods}
\subsection{Model details}

\begin{figure}[t]
 \centering
  \includegraphics[width=\linewidth]{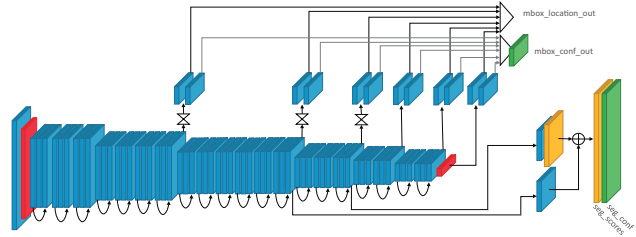}
  \vspace{-5mm}
  \caption{Our multitask network architecture for object detection and semantic segmentation.}
  \label{multinet}
\end{figure}

Extended details on our new model architecture, datasets and training strategies can be found in \cite{thesismiquel}. We select a ResNet-50 \cite{Residual} as our base network, FCN with a skip connection for the semantic segmentation task and SSD for objection detection due to their speed vs. accuracy trade-offs. Fig.~\ref{multinet} shows our multitask model.

\subsection{Datasets}

For the Pascal VOC dataset \cite{PascalRetro}, we only use the subset of samples that have ground-truth annotations for both tasks from VOC07, VOC12 and SBD\cite{SBD}.

As there is no aerial view dataset with annotations for both tasks, we interleave images from Stanford Drone Dataset \cite{StanfordDrone} for object detection and the much smaller Okutama/Swiss \cite{Laurmaa} datasets for semantic segmentation in each mini-batch.

\section{Results}
\subsection{Pascal VOC}
\begin{figure}[t]
  \centering
  \includegraphics[width=.45\linewidth]{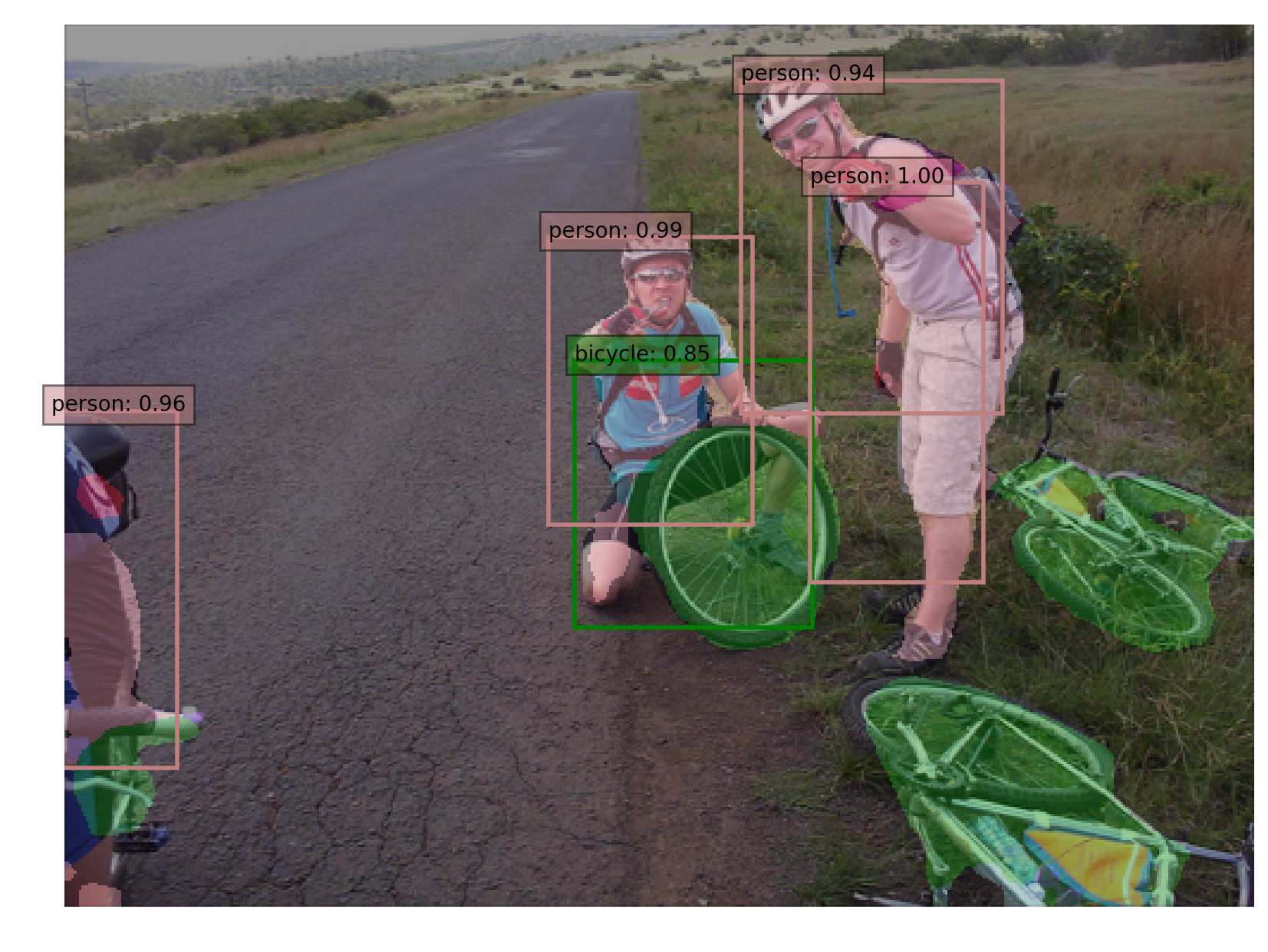}\hfill
  \includegraphics[width=.45\linewidth]{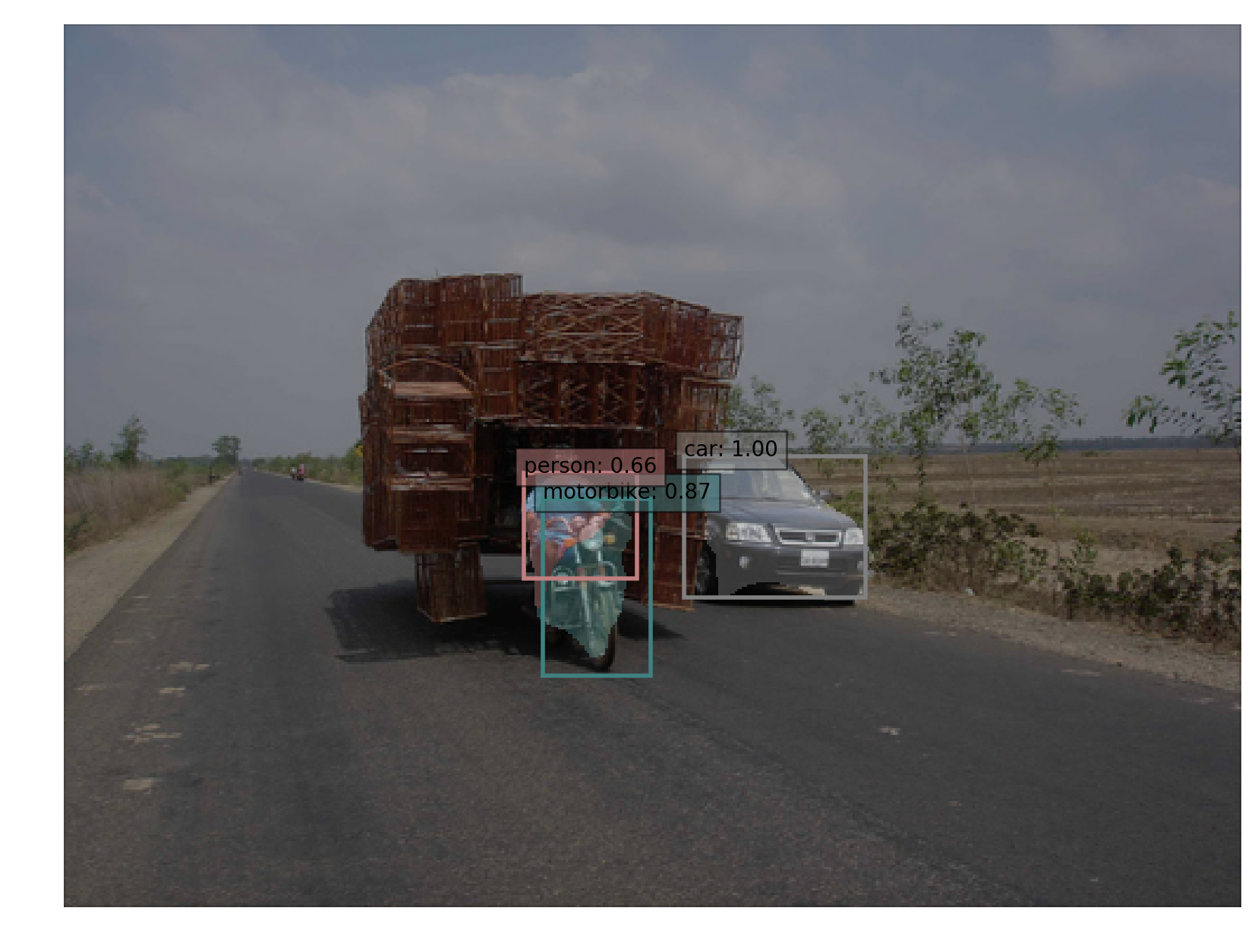}
  \vspace{-3mm}
  \caption{Pascal VOC detection and segmentation samples.}
  \label{pascalsample}
\end{figure}

We train single-task baselines with the limitations imposed by using MTL and compare those with the multitask model and reference models from the literature. Table \ref{multi-res} shows the results. Some example output images can be seen in Fig.~\ref{pascalsample}. We add color distortion for data augmentation.

\begin{table}[t]
\centering
\caption{Pascal VOC results.}
\label{multi-res}
\begin{tabular}{r|ccc|cccc}
                   & \multicolumn{3}{c}{References} \vline & \multicolumn{4}{c}{Ours} \\
                   & \cite{Laurmaa}  & \cite{SSD} & \cite{pynetbuilder} & SSD & FCN & Multi & Color \\ \hline
mIoU (\%)   &   -    &  -     &  62.5   &   -  & \textbf{56.4}  &  54.4  & 55.2 \\ 
mAP (\%)  &    74.3   & 70.4   &  - &  51.3  & - & 51.8 & \textbf{52.6}
\end{tabular}
\end{table}

\subsection{Aerial view}
\begin{figure}[t]
  \centering
  \includegraphics[width=.45\linewidth]{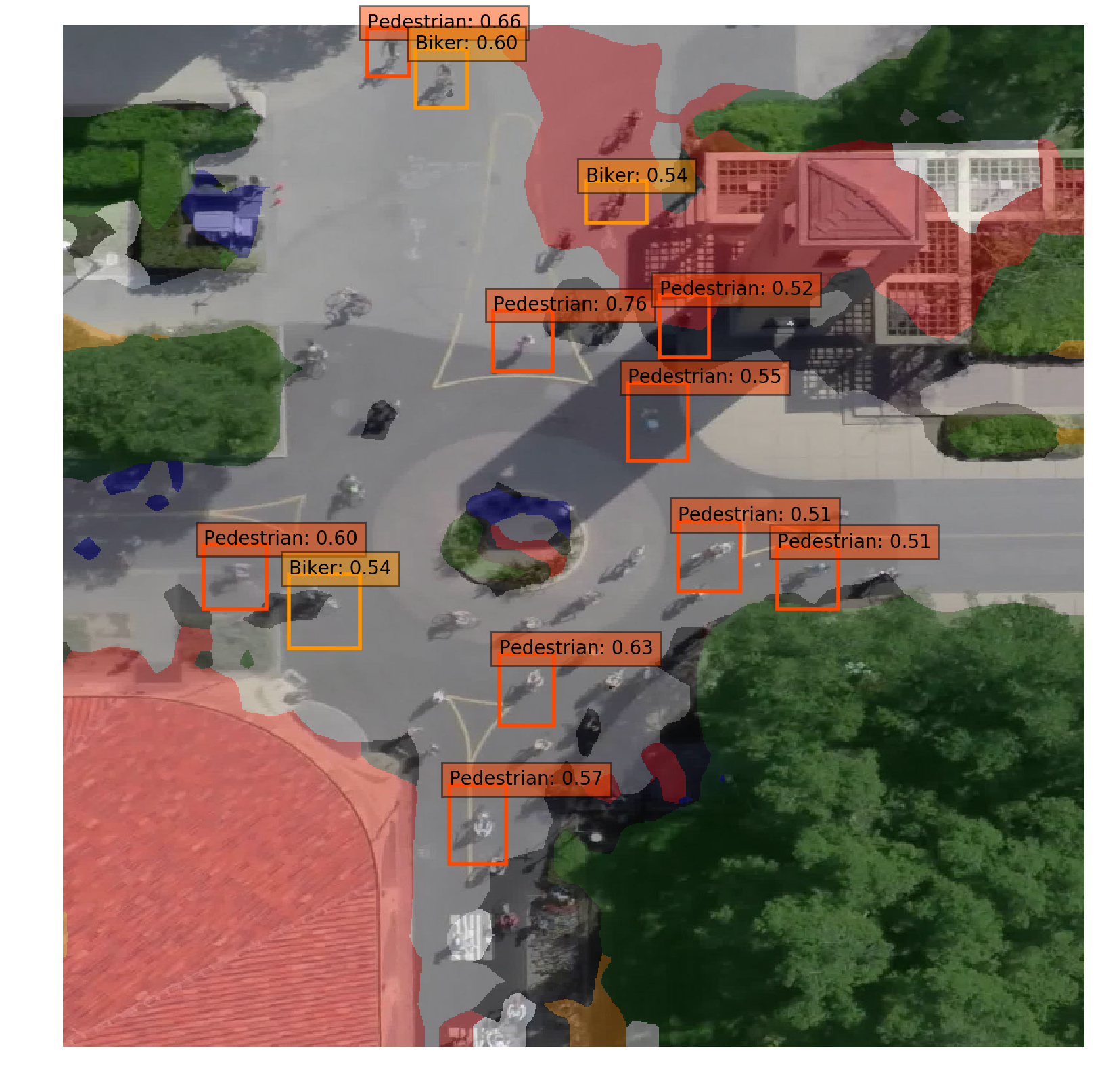}\hfill
  \includegraphics[width=.45\linewidth]{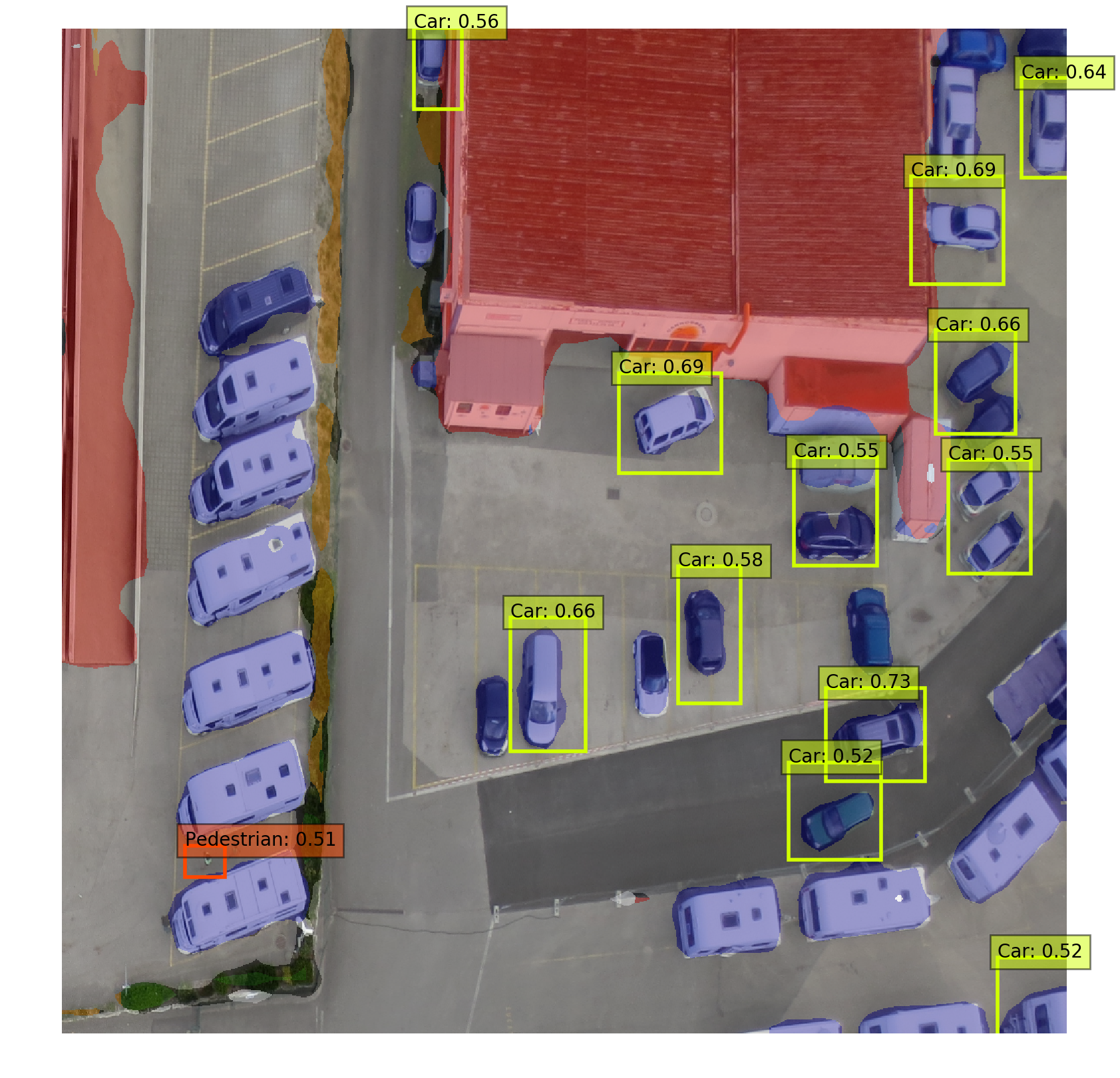}
  \vspace{-3mm}
  \caption{Aerial view detection and segmentation samples.}
  \label{aerialsample}
  \vspace{-1mm}
\end{figure}

We train single-task baselines with no limitations and compare them to our multitask model. Table \ref{aerialviewresults} shows the results in terms of accuracy and resources used when deployed on an NVIDIA GTX Titan X for the single-task models, their combination and our multitask model.  Sample output images can be seen in Fig.~\ref{aerialsample}.

\subsection{Analysis}
The compromises made due to the particularities of MTL and especially the lack of a strong data augmentation caused the final accuracy of our multitask model to lag behind that of the single-task ones trained without these although they improved in terms of speed and usage of resources, being 1.6x faster, lighter and consuming less memory than the naive solution. Compared to the single-task models trained with the same limitations, the multitask models matched or outperformed their accuracy for one of the tasks. Find a more detailed analysis in \cite{thesismiquel}.

\begin{table}[t]
\centering
\caption{Aerial view results.}
\vspace{-1mm}
\label{aerialviewresults}
\begin{tabular}{r|ccc|cc}
                   & Base & FCN & SSD & Multitask & Naive \\ \hline
mIoU (\%)   & - & 70.9    &  -        &   65.3 & \textbf{70.9}  \\ 
mAP (\%)  & -  & -   &  54.3  &  28.2 & \textbf{54.3} \\ \hline
Inf. Time (ms) & 19 & 24 & 27 & \textbf{30} & 49 \\
Memory (MB) & 1203 & 1233 & 1525 & \textbf{1552} & 2511 \\
Size (MB) & 95 & 95 & 140 & \textbf{140} & 235
\end{tabular}
\vspace{-1mm}
\end{table}

\section{Conclusion}
We conclude that MTL has the potential to give superior performance in the accuracy vs. speed over using multiple single-task models simultaneously as far as the two analyzed are concerned. This is in exchange of a training process that is more complex as it requires extra choices to be made, a new tuning of the parameters that jointly works well for each task and overcoming new challenges that were not present in the training of single-task models.

\section*{Acknowledgment}
The work was partially conducted while the first author was at Prendinger Lab, participating in the NII International Internship Program in Tokyo, Japan. 



\bibliographystyle{SSBAtrans}

\end{document}